\documentclass[10pt,twocolumn,letterpaper]{article}

\usepackage[pagenumbers]{cvpr} 

\usepackage{bm}
\usepackage{bibunits} 

\usepackage{amssymb,amsthm}
\usepackage{amsmath}
\usepackage{graphicx}
\usepackage{wrapfig,lipsum,booktabs}

\usepackage{epsfig}
\usepackage{tikz}
\usetikzlibrary{spy}
\usepackage{algpseudocode}
\usepackage{algorithm}
\usepackage{mathrsfs}

\usepackage{nicefrac}       
\usepackage{booktabs}       
\usepackage{multirow}

\usepackage{thmtools,thm-restate}
\usepackage{listings}
\usepackage{lstautogobble}  %
\usepackage{color}          %
\usepackage{zi4}            %
\definecolor{bluekeywords}{rgb}{0.13, 0.13, 1}
\definecolor{greencomments}{rgb}{0, 0.5, 0}
\definecolor{redstrings}{rgb}{0.9, 0, 0}
\definecolor{graynumbers}{rgb}{0.5, 0.5, 0.5}
\usepackage{subcaption}        
\usepackage{siunitx}               
\usepackage[export]{adjustbox}
\usepackage{makecell}
\usepackage{url}
\usepackage{tikz, pgfplots}
\usetikzlibrary{positioning}
\usepackage{capt-of}
\usepackage{diagbox}

\def\eqref#1{(\ref{#1})}

\usepackage{colortbl}
\usepackage{mdframed}

\usepackage{amsmath,amsfonts,bm}

\def\eqref#1{Eq.~\ref{#1}}

\def\1{\bm{1}}

\def\vb{{\bm{b}}}

\newcommand{\Ib}{{\bm I}}

\DeclareMathAlphabet{\mathsfit}{\encodingdefault}{\sfdefault}{m}{sl}
\SetMathAlphabet{\mathsfit}{bold}{\encodingdefault}{\sfdefault}{bx}{n}

\def\gO{{\mathcal{O}}}

\DeclareMathOperator*{\argmin}{arg\,min}

\newcommand{\xb}{{\boldsymbol x}}

\newcommand{\pb}{{\boldsymbol p}}

\newcommand{\C}{{\boldsymbol C}}
\newcommand{\e}{{\boldsymbol e}}

\newcommand{\f}{{\boldsymbol f}}

\newcommand{\p}{{\boldsymbol p}}
\newcommand{\x}{{\boldsymbol x}}
\newcommand{\y}{{\boldsymbol y}}

\newcommand{\g}{{\boldsymbol g}}

\newcommand{\epsilonb}{{\boldsymbol \epsilon}}

\newcommand{\Ed}{{\mathbb E}}
\newcommand{\Rd}{{\mathbb R}}

\newcommand{\Nc}{{\mathcal N}}
\newcommand{\Mc}{{\mathcal M}}

\newcommand{\Tc}{{\mathcal T}}

\newcommand{\code}[1] {\texttt{#1}}

\usepackage{capt-of}
\usepackage{diagbox}

\definecolor{C0}{rgb}{0.121569, 0.466667, 0.705882}
\definecolor{C1}{rgb}{1.000000, 0.498039, 0.054902}
\definecolor{C2}{rgb}{0.172549, 0.627451, 0.172549}
\definecolor{C3}{rgb}{0.839216, 0.152941, 0.156863}
\definecolor{C4}{rgb}{0.580392, 0.403922, 0.741176}
\definecolor{C5}{rgb}{0.549020, 0.337255, 0.294118}
\definecolor{C6}{rgb}{0.890196, 0.466667, 0.760784}
\definecolor{C7}{rgb}{0.498039, 0.498039, 0.498039}
\definecolor{C8}{rgb}{0.737255, 0.741176, 0.133333}
\definecolor{C9}{rgb}{0.090196, 0.745098, 0.811765}
\definecolor{trolleygrey}{rgb}{0.5, 0.5, 0.5}

\definecolor{BrickRed}{rgb}{0.6,0,0}
\definecolor{RoyalBlue}{rgb}{0,0,0.8}
\definecolor{Tdgreen}{rgb}{0,0.4,0.7}
\definecolor{pinegreen}{rgb}{0.0, 0.47, 0.44}
\definecolor{cornellred}{rgb}{0.7, 0.11, 0.11}
\definecolor{cadmiumgreen}{rgb}{0.0, 0.42, 0.24}
\definecolor{spirodiscoball}{rgb}{0.06, 0.75, 0.99}
\definecolor{mylightblue}{rgb}{0.85, 0.90, 0.94}
\definecolor{maroon}{cmyk}{0,0.87,0.68,0.32}

\usepackage{pifont}
\definecolor{cfgnull}{rgb}{0.208, 0.565, 0.953} 
\definecolor{c1}{rgb}{0.082, 0.016, 0.522}   
\definecolor{c2}{rgb}{0.349, 0.035, 0.584}   
\definecolor{c3}{rgb}{0.776, 0.165, 0.533}   
\definecolor{c4}{rgb}{0.012, 0.769, 0.631}   

\usepackage{xkcdcolors}

\newcommand{\IMG}{\textcolor{cfgnull}{IMG}}

\definecolor{cvprblue}{rgb}{0.21,0.49,0.74}
\usepackage[pagebackref,breaklinks,colorlinks,allcolors=cvprblue]{hyperref}

\def\paperID{13113} 
\def\confName{CVPR}
\def\confYear{2025}
\renewcommand{\IMG}{\textcolor{black}{FreeMCG}}
\def\eqref#1{(\ref{#1})}

\title{Derivative-Free Diffusion Manifold-Constrained Gradient for Unified XAI}

\author{Won Jun Kim$^\dagger$,\quad Hyungjin Chung$^\dagger$,\quad Jaemin Kim$^\dagger$,\quad Sangmin Lee,\quad Byeongsu Sim,\quad Jong Chul Ye\\
Korea Advanced Institute of Science and Technology\\
{\tt\small \{wonjun, hj.chung, kjm981995, leeleesang, byeongsu.s, jong.ye\}@kaist.ac.kr}\\
{\small $^\dagger$Equal contribution}
}
\begin{document}
\maketitle

\begin{abstract}
Gradient-based methods are a prototypical family of ``explainability for AI'' (XAI) techniques, especially for image-based models.
Nonetheless, they have several shortcomings in that they (1) require white-box access to models, (2) are vulnerable to adversarial attacks, and (3) produce attributions that lie off the image manifold, leading to explanations that are not actually faithful to the model and do not align well with human perception. To overcome these challenges, we introduce Derivative-Free  Diffusion Manifold-Contrained Gradients \textbf{(\IMG)}, a novel method that serves as an improved basis for explainability of a given neural network than the traditional gradient. Specifically, by leveraging ensemble Kalman filters and diffusion models, we derive a \textit{derivative-free} approximation of the model’s gradient projected onto the data manifold, requiring access only to the model’s outputs (i.e., in a completely black-box setting). We demonstrate the effectiveness of FreeMCG by applying it to both counterfactual generation and feature attribution, which have traditionally been treated as distinct tasks. Through comprehensive evaluation on both tasks - counterfactual explanation and feature attribution - we show that our method yields state-of-the-art results while preserving the essential properties expected of XAI tools.
\end{abstract}

\section{Introduction}
\label{sec:intro}

\begin{figure}
    \centering
    \vspace{-0.5cm}
    \includegraphics[width=\linewidth]{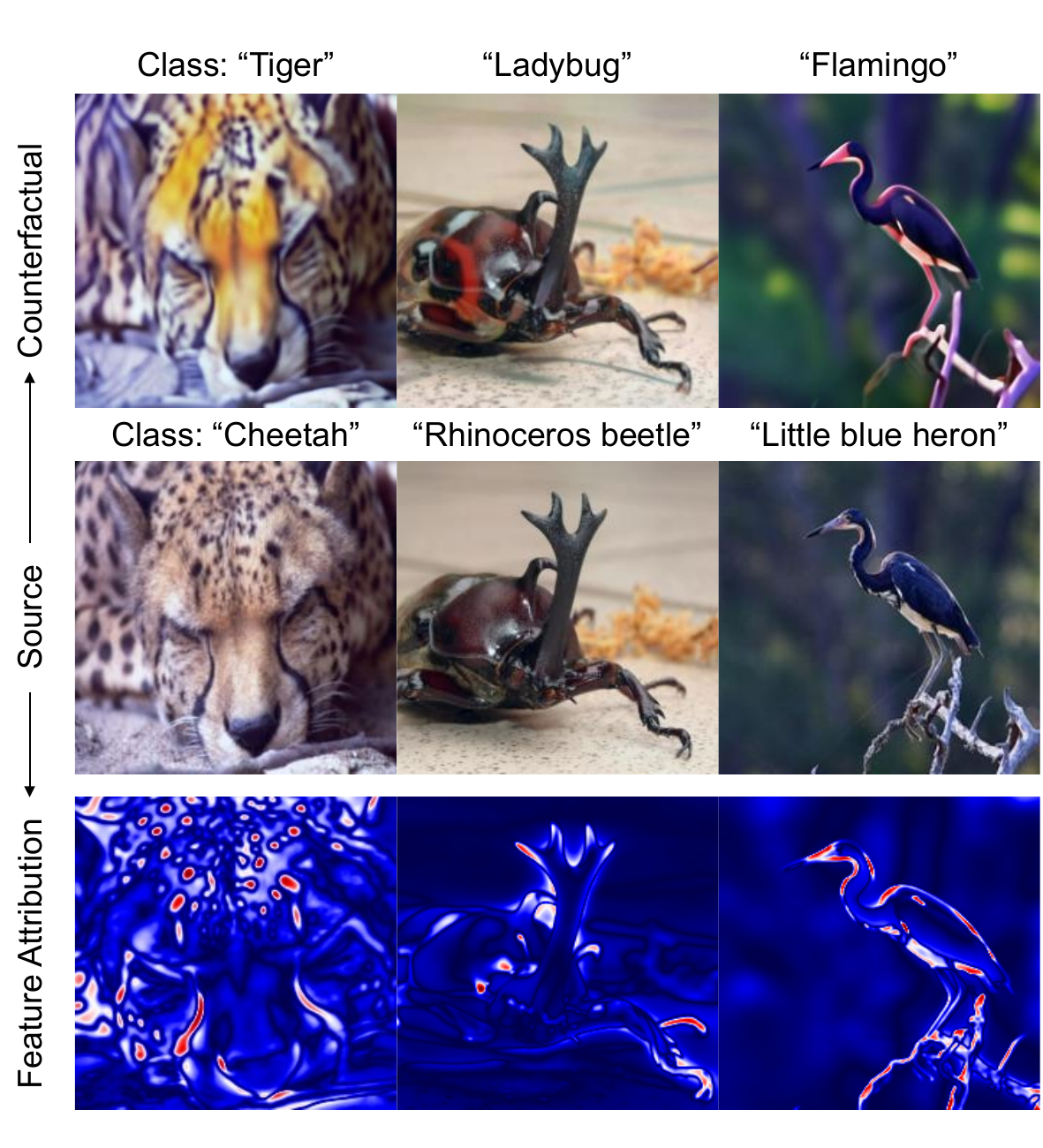}
    \vspace{-0.7cm}
    \caption{Representative results of \IMG, operating as a universal framework for black box XAI.}
    \label{fig:cover}
    \vspace{-0.5cm}
\end{figure}

\begin{figure*}
    \centering
    \includegraphics[width=0.8\textwidth]{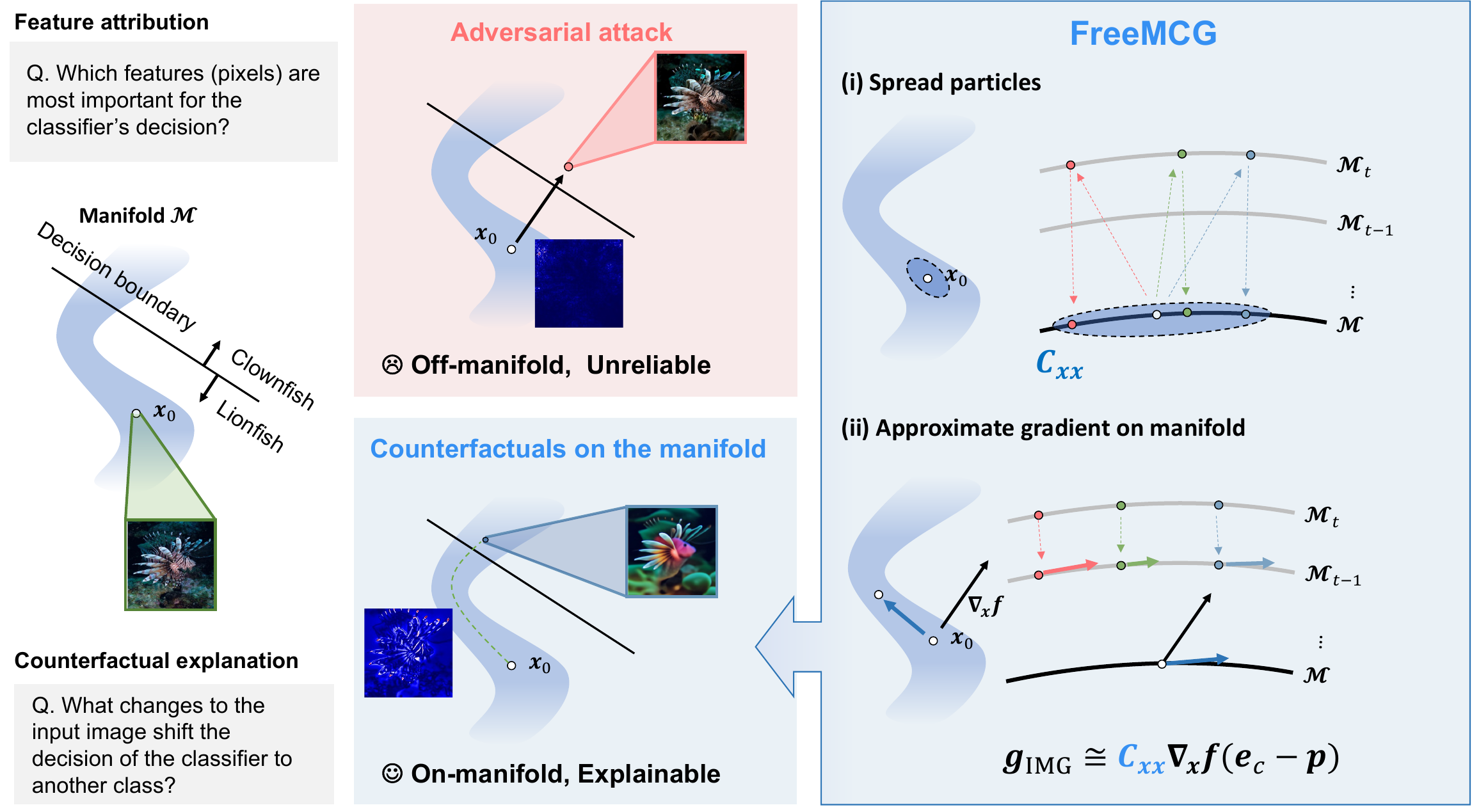}
    \caption{Unified framework of XAI for feature attribution and counterfactual explanation using our method, Derivative-Free Diffusion Manifold Constrained Gradients (\IMG). {
    Naive gradients not only violate the black box assumption, but also is unreliable and hard to interpret as they produce adversarial attacks.
    Using diffusion models to produce on-manifold ensembles, and using the covariance of these particles as the preconditioner of the gradient, FreeMCG approximates gradients on manifold, which is more reliable and interpretable.
    } }
    \label{fig:concept}
    \vspace{-0.3cm}
\end{figure*}

Understanding the decision-making process of complex deep neural networks is crucial for their deployment in high-stakes applications such as medical imaging~\cite{borys2023explainable,chen2022explainable}, autonomous systems~\cite{samek2017explainable,amodei2016concrete}, etc. Explainable AI (XAI) seeks to address this challenge by providing insights into how models arrive at specific predictions, thus enhancing trust, accountability, and usability in critical applications. Two of the most widely used tools for XAI are counterfactual explanation~\cite{verma2020counterfactual} and feature attribution\footnote{The term ``saliency maps'' are also often used interchangeably.}~\cite{lundberg2017unified}.

Among various XAI methods, the gradient of the target (e.g. classifier) model is the prototypical method of understanding the local behavior of a function around a given input data point. It has thus been used since the beginning of the rise of neural networks to explain their decisions~\citep{simonyan2013deep}. There have been many subsequent works improving the vanilla gradient to serve as better feature attributions for a deep neural network ~\citep{sundararajan2017axiomatic,selvaraju2017gradcam,ancona2018better,shrikumar2019learning,springenberg2015striving,zeiler2013visualizing}, as summarized in~\cite{wang2024gradient}. Gradients are also utilized for generating counterfactual explanations~\cite{santurkar2019image,jeanneret2023adversarial,augustin2022diffusion}, in which the gradient of the target model serves as the direction that guides the change $\bm{\delta}$ made to the original image so that prediction of the model flips to another class.

While gradient-based methods are popular explainable AI (XAI) tools, they suffer from three significant limitations:
\begin{enumerate}
    \item \textbf{Access to model gradients is not always feasible}: Obtaining gradients requires access to the model's internal weights, which is often not possible in practice. This is particularly problematic in sensitive domains like healthcare and finance, where many neural network-based services are deployed as black-box systems. Users can only interact with these systems through their outputs, without any visibility into the underlying model parameters.
    \item \textbf{Gradients provide unreliable explanations}: As shown in recent studies~\cite{adebayo2018sanity}, gradient-based methods often fail to generate meaningful explanations of model behavior. These explanations are closely related to adversarial examples~\cite{szegedy2013intriguing}, and many traditional methods resort to using models that have been trained to be adversarially robust~\cite{santurkar2019image, augustin2022diffusion}. Moreover, gradients typically do not align with the underlying data manifold, leading to explanations that are misaligned with human perception.
    \item \textbf{Lack of a unified framework for feature attribution and counterfactual explanations}: The literature on feature attribution and counterfactual explanations remains largely separate. {This is because real gradients produce adversarial attacks rather than counterfactual explanations and additional techniques are required to introduce perceptible changes. On the other hand, gradients as feature attribution also often require additional treatment to enhance their faithfulness~\cite{sundararajan2017axiomatic, shrikumar2016not}. In theory, gradients should be  usable for both applications. This raises the question of whether a single framework can effectively handle both types of explanations.} 
\end{enumerate}

To address these challenges, in this work, we propose \textit{Derivative-Free Diffusion Manifold Constrained Gradients (\IMG)}, a unified framework that can be used both for feature attribution and counterfactual explanation with only black-box access to the target model (hence without gradient computation). By leveraging diffusion models and their powerful ability to model data manifolds, we can produce on-manifold perturbations of input data points, which, combined with ideas from Ensemble Kalman Filters (EnKF)~\cite{schillings2017analysis}, can be used to approximate manifold-projected model gradients. In Fig.~\ref{fig:cover}, we show the representative results of \IMG, acting as a unified framework for XAI. Our contributions can be summarized as follows:

\begin{itemize}
    \item We present the first unified framework that encompasses both feature attribution and counterfactual explanation.
    \item We derive a novel method of estimating the gradient of the target classifier with black-box access (i.e. without access to model weights) by leveraging diffusion models and the theory of EnKF.
    \item Empirically, we demonstrate  that \IMG~achieves state-of-the-art performance, even compared to those with white-box access to the model of interest.
\end{itemize}

\section{Background \& Related Works}
\label{sec:background}

\subsection{Gradient-based XAI}
Consider the following classifier
\begin{align}
\label{eq:classifier_forward_model}
    p(y = c|\x) = \frac{\exp(\f_c(\x))}{\sum_{c'=1}^C \exp(\f_{c'}(\x))},\, \f:\Rd^d \mapsto \Rd^n,
\end{align}
where $n$ is the number of classes, and $\f(\x) := [\f_1(\x), \cdots, \f_n(\x)]$. Taking the softmax would induce the probability vector $\p = \code{softmax}(\f(\x))$. When $\x$ belongs to some class $c$, let $\code{class}(\x) = c$ be a labeling function.
The task of feature attribution for images is to find an attribution $\bm{a} \in [0, 1]^d$ that has high values in the indices where its influence to $\p$ is pronounced. On the other hand, the task of counterfactual explanation aims to find a vector $\x'$ that belongs to a target class $c'$, i.e. $\code{class}(\x') = c'$, but stays in proximity to $\x$.

\noindent\textbf{Feature attribution.}
Gradients provide a direct representation of the local behavior of a function by encapsulating how the function changes linearly in the gradient vector~\cite{simonyan2013deep}. Integrated Gradients improve upon the vanilla gradient and satisfies certain desirable properties such as sensitivity, completeness, linearity, and implementation invariance by integrating the gradients along a path from a fixed baseline (often a black image)~\cite{sundararajan2017axiomatic}. Other follow-up works have extended this approach by modifying the integration region \citep{xu2020attribution, sturmfels2020visualizing, miglani2020investigating} or altering the integrand \citep{janizek2021explaining, sikdar2021integrated, kapishnikov2021guided}, among others.

\noindent\textbf{Counterfactuals and adversarial attacks.}
Gradients obtained with respect to the input can be used to perform gradient ascent/descent to change the input towards a direction that maximizes or minimizes the model output. However, for high-dimensional input data such as images, gradient ascent or descent often yields changes that are imperceptible to humans. This is known as an \textit{adversarial attack}~\cite{goodfellow2014explaining}, because it changes the model output for a given input while the image itself remains effectively unchanged to human perception, thus having the effect of ``fooling'' a model.
The adversarial attack phenomenon occurs because gradient ascent/descent using simple gradients perturbs the image in a direction that {\em lifts} the image off the image manifold and into the ambient space~\citep{dombrowski2022diffeomorphic}.

Counterfactuals, on the other hand, should yield a {\em semantically meaningful} change to the original input $\x$ such that the target model changes its prediction. In other words, the change should be on-manifold. Naturally, to induce this on-manifold change, generative models such as GANs and diffusion models~\cite{augustin2022diffusion,jeanneret2023adversarial} have been used. However, most of these methods require access to model gradients, some even requiring that the model be adversarially robust. Moreover, these methods can be considered guided generation methods, making them inapplicable for feature attribution.

\subsection{Diffusion Models}

Diffusion model~\citep{ho2020denoising,song2020score,karras2022elucidating} is a class of generative models that sample images through a Gaussian denoising path. This path is constructed by starting from the initial data (i.e. image) distribution $p_0(\x), \x \in \Rd^d$, gradually adding Gaussian noise, and reaching approximately the Gaussian distribution $p_T(\x) \approx \Nc(\bm{0}, \Ib)$ at $t = T$. For the choice of the Gaussian perturbation $p(\x_t|\x_0) = \mathcal{N}(\x_0, \sigma_t^2\Ib)$, the reverse generation trajectory reads
\begin{align}
\label{eq:pfode}
    d\x_t = -\sigma_t\nabla_{\x_t} \log p(\x_t)\,dt = \frac{\x_t - \mathbb{E}[\x_0|\x_t]}{\sigma_t}\,dt,
\end{align}
where the initial point is sampled from the reference distribution $\x_T \sim p_T(\x_T)$. Here, Tweedie's formula~\citep{efron2011tweedie} $\Ed[\x_0|\x_t] = \x_t + \sigma_t^2\nabla_{\x_t} \log p(\x_t)$ is used for the second equality. Diffusion models are trained as denoisers that estimate this posterior mean
\begin{align}
\label{eq:dsm}
    \theta^* = \argmin_\theta \Ed_{t, \x_t, \x_0 }\left[\|D_\theta(\x_t) - \x_0\|_2^2\right],
\end{align}
which can be plugged into \eqref{eq:pfode} as $\Ed[\x_0|\x_t] \approx D_{\theta^*}(\x_t)$ at test time.

In \citet{chung2022improving}, assuming that $\Mc$ is a locally linear subspace, it was further shown that the distribution of {\em noisy} data $p(\x_t)$ is then concentrated on shell-like manifold $\Mc_t$. Moreover, one can show that Tweedie's formula is an orthogonal projection from $\Mc_t$ to $\Mc$. In other words, when the goal is to produce samples on the manifold $\Mc$ in a cost-effective way, one can use Tweedie's formula to generate samples in one step. This matches our intuition up to moderate values of $t$, where the posterior mean gives us a relatively good approximation of the real samples.

\paragraph{Guided sampling.}
Often, it is desirable to sample from the posterior $p(\x|\y)$ instead of from the prior. A standard approach to achieve this is to use guided sampling by Bayes decomposition
\begin{align}
    \nabla_{\x_t} \log p(\x_t|\y) = \nabla_{\x_t} \log p(\x_t) + \nabla_{\x_t} \log p(\y|\x_t), 
\end{align}
where the gradient of the log-likelihood term measures the {\em fidelity} with respect to the condition $\y$. One can either train a separate neural network (e.g. classifier~\cite{dhariwal2021diffusion,song2020score}) to estimate the term or use some form of approximation~\cite{chung2023diffusion,song2023pseudoinverseguided}. Sampling from the posterior then boils down to iterating the denoising steps of the unconditional PF-ODE in \eqref{eq:pfode} and the gradient step that maximizes fidelity.
However, regardless of the forward model\footnote{For instance, in image restoration, the degradation operator is the forward model. In classification, the neural network classifier is the forward model, as in \eqref{eq:classifier_forward_model}.}, most of the methods require access to the {\em gradient} of the forward model, which is often hard to compute due to reasons including safety and complexity.

\subsection{Gradient-free optimization}

When one can only evaluate $\f$ but cannot compute its gradient, gradient-free (zeroth-order) optimization~\cite{liu2020primer}, where a single or multiple set of samples can be used to calculate the finite difference, alongside techniques such as Gaussian smoothing~\cite{nesterov2017random} can be used to estimate the true gradient. Ensemble Kalman Filters (EnKF)~\cite{schillings2017analysis}, in terms of inverse problem solving, is also a widely used method for derivative-free optimization, where the ensemble of particles and its covariance information is used as a proxy for gradient information. Recently, the use of EnKF was extended to inverse problem-solving with diffusion models~\cite{zheng2024ensemble}, showing that the technique can be readily used for diffusion guidance. 
In the proposed method, we leverage EnKF to approximate the unknown gradient.

\section{Main Contribution: \IMG}
\label{sec:main}

\subsection{Derivation of \IMG}
From \eqref{eq:classifier_forward_model}, one can derive that
\begin{align}
    \frac{\partial}{\partial \f_{c'}} \log p(y=c|\f_{c'}) = \delta_{cc'} - p(c'|\f_{c'}),
\label{eq:cp1}
\end{align}
where $\delta_{cc'}$ is the Kronecker delta function which evaluates to 1 if and only if $c' = c$. Vector form of \eqref{eq:cp1} reads
\begin{align}
    \nabla_{\f} \log p(y=c|\f) = \e_c - \p,
\label{eq:vcp1}
\end{align}
where $\e_c$ is a one-hot vector with one at index $c$, and $\p \in \Rd^n$ is the classifier probability. From \eqref{eq:vcp1}, we further have that
\begin{align}
    \nabla_{\x} \log p(y=c|\x) = \left(
    \frac{\partial \f(\x)}{\partial \x}
    \right)^{\!\top} \!\!(\e_c - \p(\x)),
\label{eq:goal}
\end{align}
where we wrote $\p(\x)$ to emphasize that the probability vector is a function of $\x$.
To leverage the ideas from EnKF, we are interested in the pre-conditioned version of the gradient, where the covariance of the particles is used
\begin{align}
 \g_{\rm Free}:=   \C_{\x\x} \nabla_{\x} \log p(c|\x)
  = \C_{\x\x} \left(
    \frac{\partial \f(\x)}{\partial \x}
    \right)^{\!\top}\!\!(\e_c - \p(\x)),
    \label{eq:goal2}
\end{align}
where $\C_{\x\x} := \frac{1}{K}\sum_{k=1}^K (\x^{(k)} - \bar\x)(\x^{(k)} - \bar\x)^\top$, and we use $\bar\x$ to denote the sample mean $\Ed[\x^{(k)}]$. 

\paragraph{\IMG~produces on-manifold gradients.}

\begin{figure}
    \centering
    \includegraphics[width=\linewidth]{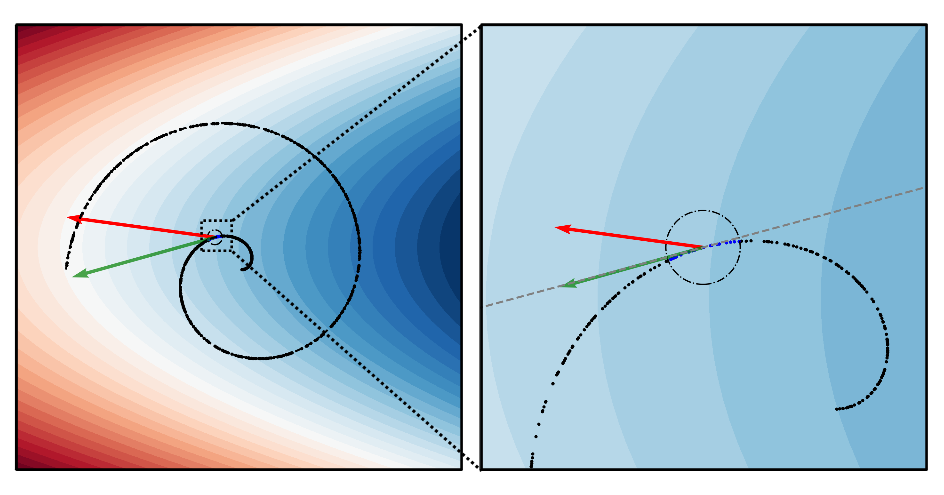}
    \caption{Toy example of \IMG, illustrating the gradient (red arrow) and the gradient aligned with the tangent of manifold (gray dashed) approximated by \eqref{eq:img_approx} (green arrow), using neighbor data points (blue dots).}
    \label{fig:img-onmanifold}
    \vspace{-0.3cm}
\end{figure}

One may question the need for preconditioning the gradient with the empirical covariance matrix. In Theorem~\ref{thm:proj}, we show that this preconditioning yields gradients that are on the data manifold $\Mc$, producing results that are well-aligned with human perception, and robust to adversarial attacks. The proof can be found in Appendix~\ref{sec:proof}.

\begin{restatable}{theorem}{proj}
    Suppose that $\Mc$ is locally linear at $\x$, meaning that in the neighborhood of $\x$, the manifold aligns with its tangent space $\Tc_\x\Mc$. Suppose further, that the particles $\x^{(k)} \in \Mc$. Then, $\g_{\rm Free}$ approximately lies within $\Mc$. Precisely, $\C_{\x\x}$ acts as a linear transformation, expanding along $\Tc_\x \Mc$ and contracting the vectors normal to $\Tc_\x \Mc$.
\label{thm:proj}
\end{restatable}

Theorem~\ref{thm:proj} states that when the particles $\x^{(k)}$ lies on the manifold surrounding the original data point $\x$, 
$\C_{\x\x}$ captures the geometric properties of the tangent space $T_\x \Mc$.
Specifically, the covariance matrix refines the vanilla gradient by contracting direction that lead off the manifold, which helps prevent the possibility of adversarial examples. Additionally, it expands along the direction of greater variability among the particles, reducing the likelihood of converging to a single original point.

\paragraph{EnKF for derivative-free gradient}
However, notice that $\C_{\x\x}\left(\frac{\partial \f(\x)}{\partial \x}\right)^\top$ is hard to compute, as we do not have access to the gradient of $\f$. Thus, we invoke the following.

\begin{restatable}{theorem}{imgenkfthm}
    Suppose $\|\nabla_{\x}^2 \f(\x)\|$ be uniformly bounded and there exists $\delta>0$ such that $\x^{(k)} \in \{\x \in \Rd^d : \| \x - \bar\x \| \le \delta \}$ for all $k$. Then, the following approximation holds
    \begin{align}
    \label{eq:img_approx}
        \C_{\x\f} = \C_{\x\x}\nabla_{\x}\f(\x)^\top + \gO(\delta^3),
    \end{align}
    where $\C_{\x\f} := \frac{1}{K}\sum_{k=1}^K (\x^{(k)} - \bar\x)(\f(\x^{(k)}) - \bar\f)^\top$ and $\bar\f := \frac1K \sum_{k=1}^K \f(\x^{(k)})$.
    \label{prop:ensemble_kalman}
\end{restatable}

The proof along with a formal treatment on approximation errors can be found in \cref{sec:proof}. Apply \cref{prop:ensemble_kalman} to \eqref{eq:goal2}, we have
\begin{align}
    &\C_{\x\x} \nabla_{\x} \log p(y=c|\x) \approx \C_{\x\f}(\e_c - \p) \notag \\
    &= \frac{1}{K}\sum_{k=1}^K (\x^{(k)} - \bar\x)(\f(\x^{(k)}) - \bar\f)^\top(\e_c - \p(\x)),
\label{eq:img_base}
\end{align}
given the particles $\x^{(k)}$. We provide the pseudocode for computing \IMG~in Alg.~\ref{alg:img}.

Although we now have the machinery to compute the approximate gradient, notice that we have to choose the position of the particles $\x^{(k)}$. For this, three conditions must be met. The particles should be 1) near $\x$, 2) on-manifold, and 3) fast to generate. The first two conditions are met by perturbing $\x$ to obtain forward-diffused samples $\x_t^{(k)}$, and running reverse PF-ODE to all the particles, as used in~\cite{meng2021sdedit,zheng2024ensemble}. However, since this would be prohibitively slow, we simply choose to use the Tweedie denoised estimates, which we denote as $\x_{0|t}^{(k)}$. In short, our gradient reads
\begin{align}
\label{eq:img}
    \g_{\rm Free} := \Ed\left[
    (\x_{0|t}^{(k)} - \bar\x_{0|t}^{(k)})(\f(\x_{0|t}^{(k)}) - \bar\f(\x_{0|t}^{(k)}))^\top (\e_c - \p)
    \right],
\end{align}
where the expectation is over the diffusion timesteps $t$ and the number of particles $k$. 
Notice that 
under our assumption of local linearity, we can use Proposition 2 of \cite{chung2022improving}, which states that Tweedie's formula~\cite{efron2011tweedie} is locally an orthogonal projection to the manifold $\Mc$, satisfying the assumption of Thm.~\ref{thm:proj}.

We illustrate a 2D toy example in Fig. \ref{fig:img-onmanifold}. Considering $f = -x+y^2$ with its gradient $\nabla_{\x} \f(\x) = (-1,2y)$, we sample particles around the point $\x$ and calculate the covariance matrix. Then, applying \eqref{eq:img_approx}, we obtain a gradient aligned with the direction of the manifold without directly computing the gradient of the $f$. In our main experiments, we show that \IMG~indeed produces a gradient that produces on-manifold gradients so that it avoids being an adversarial attack.
In the following, we elaborate on practical implementations of \eqref{eq:img} for feature attribution and counterfactual explanation, as well as their intuition, respectively.

\subsection{Applications of \IMG}
\label{sec:img_applications}

\begin{algorithm}[!t]
\caption{\IMG~Gradient Computation}
\begin{algorithmic}[1]
    \Function{\IMG}{$\f$, $\{\x^{(k)}\}_{k=1}^K$, $\e_c$, $\pb$}
        \State $\bar{\x} \gets \frac{1}{K} \sum_{k=1}^K \x^{(k)}$
        \State $\bar{\f} \gets \frac{1}{K} \sum_{k=1}^K \f(\x^{(k)})$
        \For{$k=1$ to $K$}
            \State $\bm{\Delta}^{(k)} \gets (\x^{(k)} - \bar{\x}) (\f(\x^{(k)}) - \bar{\f})^\top (\e_c - \pb)$
        \EndFor
        \State \Return $\g_{\rm Free} \gets \frac{1}{K} \sum_{k=1}^K \bm{\Delta}^{(k)}$
    \EndFunction
\end{algorithmic}\label{alg:img}
\end{algorithm}
\begin{algorithm}[!t]
\caption{\IMG~for feature attribution}
\begin{algorithmic}[1]
    \Require Classifier $\f$, Diffusion model $\epsilonb_\theta$, Input $\x$, Number of particles $K$, Target class vector $\e_c$, Diffusion timestep range $[l, u]$
    \State $\p = \mathrm{softmax}(\f(\x))$
    \State $\x_0 \gets \x$
    \For{$k=1$ to $K$} \Comment{$K$ ensemble particles}
        \State $t \sim \mathcal{U}[l, u]$ \Comment{Random timestep for each particle}
        \State $\x_t^{(k)} \sim p(\x_t|\x_0)$ \Comment{Forward diffuse}
        \State $\x_{0|t}^{(k)} \gets (\x_t - \sqrt{1 - \bar\alpha_t}\epsilonb_\theta(\x_t))/\sqrt{\bar\alpha_t}$
        \Statex \Comment{Tweedie denoising}
    \EndFor
    \State $\g_{\rm Free} \gets$ \Call{\IMG}{$\f$, $\epsilonb_\theta$, $\x$, $\{\x_{0|t}^{(k)}\}_{k=1}^K$, $\e_c$, $\pb$}
    \State \Return $\g_{\rm Free}$
\end{algorithmic}\label{alg:img_att}
\end{algorithm}
\begin{algorithm}[!t]
\caption{\IMG~for CE with reverse diffusion}
\begin{algorithmic}[1]
    \Require Classifier $\f$, Diffusion model $\epsilonb_\theta$, Input $\x$, Number of particles $K$, Target class vector $\e_{c'}$, Diffusion params $t', \tilde\beta_t, \bar\alpha_t, \eta, \gamma_t$
    \State $\x_0 \gets \x$
    \State $[\x_{t'}^{(k)}] \sim p(\x_{t'}|\x_0)$ 
    \Comment{$[\cdot]$ denotes $K$ batched particles}
    \For{$t=t'$ to $1$} \Comment{reverse diffusion}
        \State $[\hat\epsilonb_t^{(k)}] \gets \epsilonb_\theta([\x_t^{(k)}])$
        \State $[\x_{0|t}^{(k)}] \gets ([\xb_{t}^{(k)}] - \sqrt{1 - \bar\alpha_t}[\hat\epsilonb_t^{(k)}]) / \sqrt{\bar\alpha_t} $
        \Statex \Comment{Tweedie denoising}
        \State $[\p^{(k)}] \gets \mathrm{softmax}(\f([\x_k]))$
        \State $[\g_{\rm Free}^{(k)}] \gets$ \Call{\IMG}{$\f$, $[\x_{0|t}^{(k)}]$, $\e_{c'}$, $\pb^{(k)}$}
        \State $[\g^{(k)}] \gets \alpha [\g_{\rm Free}^{(k)}] + \beta(\x - [\x_{0|t}^{(k)}])$
        \State $[\epsilonb^{(k)}] \sim \Nc(0, \Ib)$
        \State $[\x_{t-1}] \gets \sqrt{\bar\alpha_{t-1}}[\x_{0|t}]^{(k)} + \sqrt{1 - \bar\alpha_{t-1} - \eta^2 \tilde\beta_t^2}[\hat\epsilonb_t]$
        \Statex \hspace{2em} $+ \, \eta \tilde\beta_t[\epsilonb^{(k)}] + \gamma_t[\g^{(k)}]$
    \EndFor
    \State \Return $\x^{\mathrm{(CE)}} = \frac{1}{K}\sum_{k=1}^K\x_0$ \Comment{Average of particles}
\end{algorithmic}\label{alg:img_ce_rev_diff}
\end{algorithm}

\noindent\textbf{\IMG~for feature attribution.}
This corresponds to the case where the target class $c'$ aligns with the prediction of the classifier $c$. Notice that $\f(\x^{(k)}) - \bar\f \in \Rd^n$ where $\f_c$ indicates the average drop in logit value after the Tweedie perturbation. Moreover, when $c' = c$, the only positive entry in $\e_c - \p$ will be at the index $c$, positively weighting the part that changes the predicted class label, while negatively weighting the part that is irrelevant. As a concrete example, consider the following 4-class example
\begin{align}
\label{eq:example_fa}
    &\e_c = [1, 0, 0, 0]^\top, \p = [0.7, 0.1, 0.1, 0.1]^\top \notag \\
    &\e_c - \p = [0.3, -0.1, -0.1, -0.1]^\top,
\end{align}
where only the perturbations that dropped the probability of the first class will be positively weighted, while for the other indices, there will be an opposite force.

For implementation, we found that using multiple levels of $t$ was crucial for encapsulating the attribution that acts both for coarse and fine scales. Similar to how score distillation sampling (SDS)~\cite{poole2022dreamfusion} is recently used~\cite{siglidis2024diffusion}, we constrain the time range to $t \in [0.1, 0.7]$. For visualization purposes, we take the mean across the color channels and plot the absolute value. The algorithm for feature attribution using \IMG~is presented in Alg.~\ref{alg:img_att}.

\begin{figure*}
    \centering
    \includegraphics[width=0.96\linewidth]{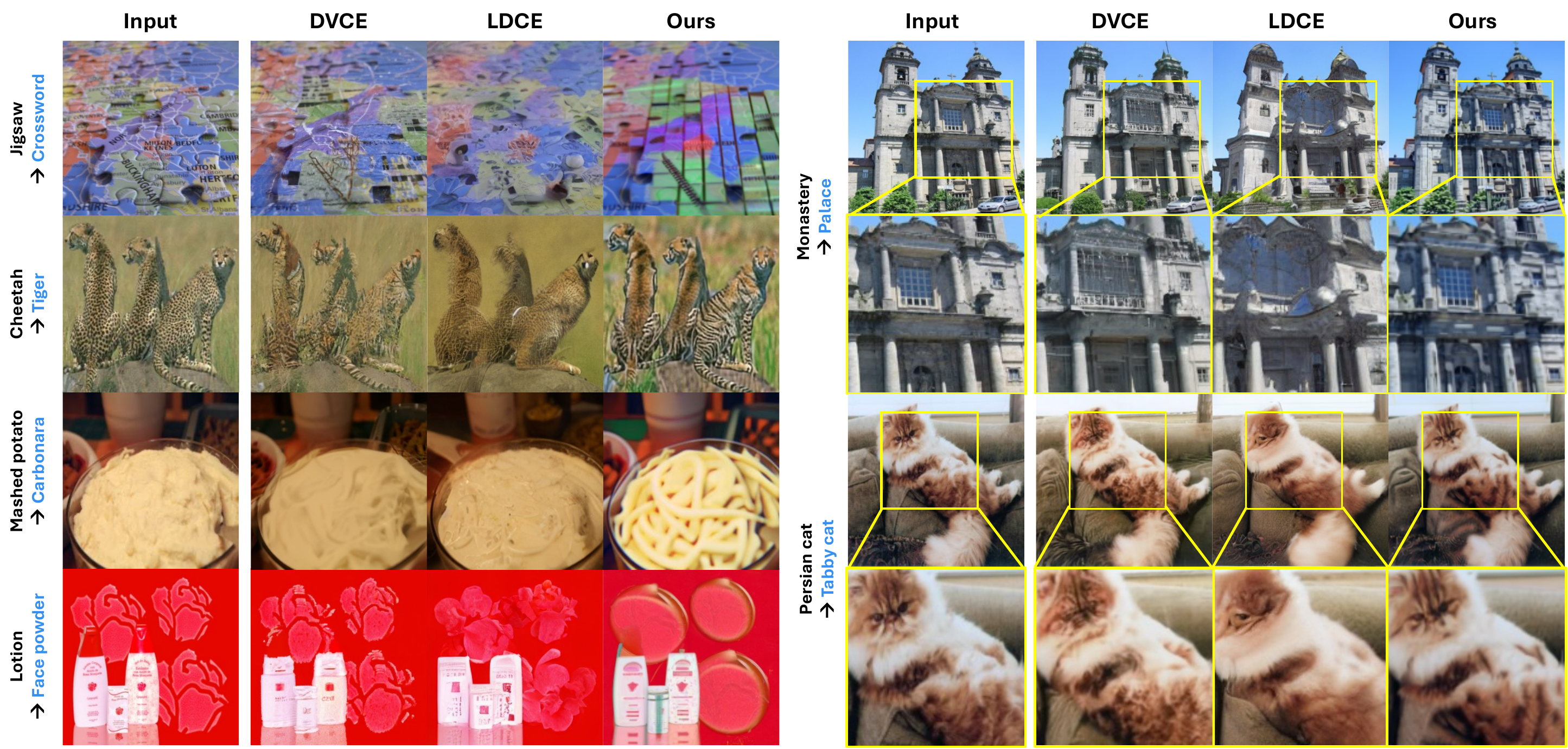}
    \caption{\textbf{Counterfactual Generation.} Images on the leftmost column are given as input, and the counterfactual for the target class (shown in blue) are generated using each method. Baseline methods generate counterfactuals by perturbing the image towards the direction of a given classification neural network's gradients. Not only are these methods not applicable to situations without access to model weights, gradients almost always point away from the image manifold and into the ambient space, leading to imperceptible changes (i.e., adversarial attack) or changes that produce unnatural, off-manifold, images. FreeMCG on the other hand produces semantically meaningful changes while keeping the image on-manifold.}
    \label{fig:imagenet-cf}
    \vspace{-0.3cm}
\end{figure*}

\noindent\textbf{\IMG~for counterfactual explanation.}
While \eqref{eq:img} can directly serve as the attribution, for counterfactual explanation, the gradient should be used as the direction of ascent toward a counterfactual sample. 
Here, we can iteratively apply gradient ascent with FreeMCG, i.e.
\begin{align}
\label{eq:img_ce}
    \x_{i+1} = \x_i + \alpha\g_{\rm Free}(\x_i) + \beta (\x - \x_i),
\end{align}
where $\alpha$ is the step size weighting the \IMG, and $\beta$ weights the proximal regularization to the initial point $\x$. Notice that the implementation of \IMG~for counterfactual explanation differs from that of feature attribution, as we have a different target class $c'$ from $c$. For instance, consider the example in \eqref{eq:example_fa}, but the target class is now 2. Then, in the earlier stage of gradient ascent, we have
\begin{align}
\label{eq:example_cf}
    &\e_c = [0, 1, 0, 0]^\top, \p = [0.7, 0.1, 0.1, 0.1]^\top \notag \\
    &\e_c - \p = [-0.7, 0.9, -0.1, -0.1]^\top.
\end{align}
In other words, the components correlated with the original class $c$ is strongly negatively pulled down, whereas the components correlated with the target class $c'$ are strongly positively pushed up.

\noindent\textbf{\IMG~meets reverse diffusion.}
While iterative gradient ascents using \IMG~in \eqref{eq:img_ce} gives us a fast way to generate counterfactual explanations, we find that this process is sensitive to hyperparameters and produces suboptimal results. To fully leverage the generative nature of diffusion models, we instead propose a variant that utilizes reverse diffusion, as is done in other methods that use diffusion models for counterfactual explanation~\cite {jeanneret2022diffusion,augustin2022diffusion,farid2023latent}. Specifically, we use an SDEdit~\cite{meng2021sdedit}-like approach, where we forward diffuse our initial sample up to time $t'$ and run reverse guided sampling in~\eqref{eq:img_ce} to the Tweedie denoised estimates. The algorithm is presented in Alg.~\ref{alg:img_ce_rev_diff}. Notice that we do not incorporate the Jacobian of $\epsilonb_\theta$, which is typically done in methods like~\cite{chung2023diffusion,song2023pseudoinverseguided}, and simply resort to the gradients with respect to the Tweedie estimates~\cite{chung2024decomposed} to save computation. For further details, see Appendix~\ref{sec:implementation_details}.
This approach may be slower as it requires reverse diffusion, but produces higher-fidelity results (Fig.~\ref{fig:imagenet-cf}). 
Therefore, all experiments on counterfactual explanation with \IMG~use this reverse diffusion approach unless specified otherwise.

\section{Experiments}
We demonstrate our method from both feature attribution and counterfactual perspectives. To evaluate generalization and domain-specific performance, we compare our method on ImageNet~\cite{deng2009imagenet} and MIMIC CXR~\cite{johnson2019mimic} dataset, using images with a resolution of 256$\times$256. We use regular pretrained classifier models with the ResNet-50~\cite{he2016deep} and unconditional diffusion models~\cite{dhariwal2021diffusion} trained on each dataset. Experimental details are provided in Appendix~\ref{sec:experiment}.



\subsection{Evaluation protocol}\label{subsec:eval-protoc}
\noindent\textbf{Counterfactuals.} An ideal counterfactual makes semantically meaningful, human-interpretable yet minimal changes to the input image so that it is classified as the target class by the model of interest. This allows the user to gain visual insight into the decision boundary of the model. As counterfactual generation aims to create an example at a point on or near the decision boundary closest to the input image (Fig~\ref{fig:concept}), the final generated counterfactual must be anchored on the original input image and should not be an unconditionally generated image of the target class.
Therefore, evaluation of counterfactuals involves the following inter-related criteria:

\begin{enumerate}
    \item \textbf{Meaningful change towards target class}
    \begin{enumerate}
        \item \textbf{Flip rate:} Is the generated counterfactual classified as the intended target class by the model of interest?
        \item \textbf{Perceived change:} Does the generated counterfactual contain semantically meaningful changes to a human perceiver? (i.e. can it be distinguished from an \textit{adversarial attack}?)
    \end{enumerate}
    \item \textbf{Similarity to the input}
    \begin{enumerate}
        \item \textbf{L2 distance:} Are generated counterfactuals close to the input images in terms of distance?
        \item \textbf{Perceived similarity:} Do generated counterfactuals and input appear similar to a human perceiver?
    \end{enumerate}
    \item \textbf{Realism of generated image}
    \begin{enumerate}
        \item \textbf{Frechet Inception Distance (FID)~\cite{heusel2017gans}:} Is the distribution of the generated counterfactual images similar to distribution of original input images?
        \item \textbf{Perceived realism:} Do the images appear realistic to a human perceiver?
    \end{enumerate}
\end{enumerate}

A crucial aspect about these metrics is that they must be read and understood in concert, rather than disparately as separate measures. In particular, note that an \textit{adversarial attack}~\cite{goodfellow2014explaining, costa2024adversarial}, which flips the decision of the classification neural network but does not introduce changes to the input image that are meaningful to a human perceiver (which are meaningless as explainability tools), can speciously score well on aspects such as similarity or realism. Assessment of similarity and realism scores must thus be pre\textbf{}dicated on the fact that there are meaningful changes that have been made towards the target class and viewed in proportion to that change: \code{Change} $\times$ (\code{Similarity} + \code{Realism}).

{Flip rate, L2 distance, and FID are computed using 1K generated counterfactuals (one from each class of the ImageNet validation set). Following the experimentation scheme in LDCE~\cite{farid2023latent}, we select the category closest to the input image based on the WordNet~\cite{miller1994wordnet} hierarchy as the target class for counterfactual generation. Perceived change, perceived similarity, and perceived realism were measured through a human user study (Appendix~\ref{sec:human-user-study}). }Results are provided in Table~\ref{tab:counterfactual-eval}.

\begin{figure*}
    \centering
    \includegraphics[width=0.95\linewidth]{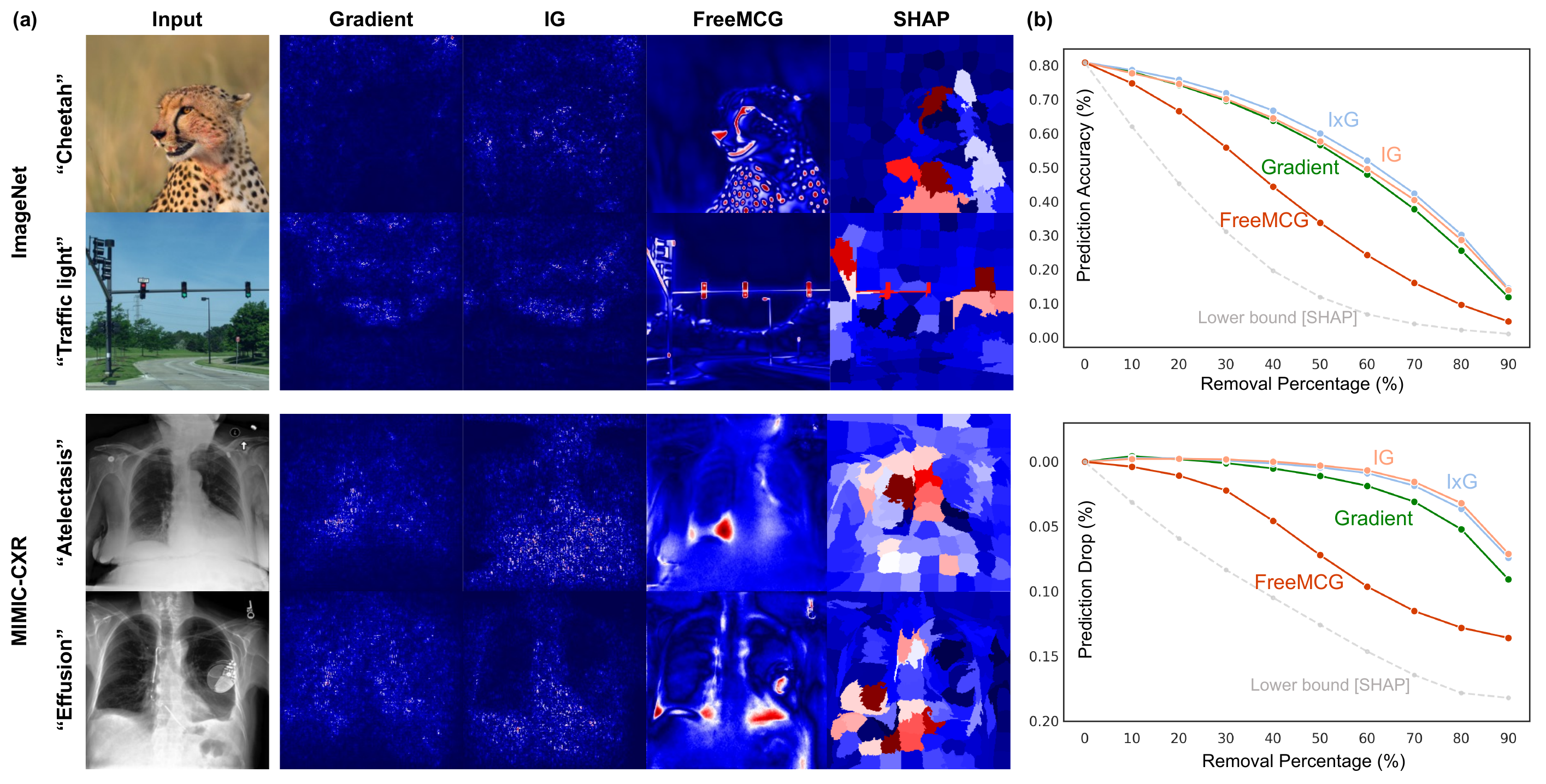}
    \caption{(a) Feature attribution maps obtained using each method, (b) Corresponding ROAD~\cite{rong2022consistent} evaluation results. Steeper decline is better, and SHAP is given as reference for the lower bound of this decline. For ImageNet, ROAD is calculated as described in~\cite{rong2022consistent}. For MIMIC chest X-ray data, results for decline in prediction accuracy was noisy due to the lower starting accuracy of the classifier, low number of classes, and severe data imbalance between classes; therefore, we instead measure the drop in prediction for the originally predicted class. The principle remains the same in that features (pixels) are removed in order of purposed importance from each feature attribution map to observe the change in classifier predictions.}
    \label{fig:smap}
    \vspace{0.2cm}
\end{figure*}

We select DVCE~\cite{augustin2022diffusion} and LDCE~\cite{farid2023latent} as baselines, leveraging diffusion models to generate counterfactuals. Note that these baseline methods require access to the model weights while \IMG~does not. In addition, while the proposed method in DVCE includes a version with cone projection of model weights to the weights of an adversarially robust classifier, we use only a regular non-robust classifier for comparison as we are assuming lack of access to the training data or adversarially robust weights.


\noindent\textbf{Feature attribution.} 
Feature attribution maps can be evaluated by observing the change in model predictions whilst removing the pixels in order of purported importance from the attribution.
We compare our approach with existing gradient-based attribution methods, including the vanilla gradient~\cite{simonyan2013deep}, Integrated Gradients~\cite{sikdar2021integrated} and InputXGradient~\cite{shrikumar2016not}.  
{Shapley additive explanation (SHAP)~\cite{lundberg2017unified} is an effective model-agnostic feature attribution method that determines the importance of each feature by assessing the change in classifier predictions after removing a certain feature from possible combinations of features that make up a data point. For images, this amounts to dividing an image into clusters of pixels (`superpixels') and determining the importance of each superpixel by observing the effect of their removal. Since the size of the superpixels in the SHAP is usually very large, the SHAP maps are not usually as perceptually informative as gradient-based attribution methods.

As for the metric for the feature attribution,
Remove and Debias (ROAD)~\cite{rong2022consistent} is a removal-based evaluation metric that minimizes information leakage from the shape of the pixel removal mask using noisy linear imputation. 
As such, ROAD is ideally suited with SHAP method, so under the assessment by removal-based metrics such as ROAD, SHAP feature attributions (Fig.~\ref{fig:smap})  can be used as a sort of `gold standard' to determine which attributions are most faithful to the model being explained.

\subsection{Results}

\begin{table*}[!thb]
    \centering
    \resizebox{0.8\textwidth}{!}{
    \begin{tabular}{lcccccc}
    \toprule
    & \multicolumn{2}{c}{\textbf{Change}} & \multicolumn{2}{c}{\textbf{Similarity}} & \multicolumn{2}{c}{\textbf{Realism}} \\
    \cmidrule(lr){2-3} \cmidrule(lr){4-5} \cmidrule(lr){6-7} 
      & Flip Rate $\uparrow$ & Perceived change $\uparrow$ & L2 $\downarrow$ & Perceived similarity $\uparrow$ & FID $\downarrow$ & Perceived realism $\uparrow$ \\
    \midrule
    DVCE~\cite{augustin2022diffusion} & \textbf{85.8} & 2.38 & \textbf{995.92} & 3.26 & 71.38 & 2.77 \\
    LDCE~\cite{farid2023latent} & 61.7 & 2.59 & 1351.39 & 2.87 & \textbf{63.04} & 2.59 \\
    Ours & 51.5 & \textbf{3.74} & 1001.02 & \textbf{3.52} & 66.89 & \textbf{3.13} \\
    \bottomrule
    \end{tabular}
    }
    \caption{Quantitative (Flip Rate, L2, FID) and human user study (perceived change, perceived similarity, perceived realism) results.}
    \label{tab:counterfactual-eval}
    \vspace{-0.4cm}
\end{table*}

\noindent\textbf{Counterfactual generation.}
Counterfactual generation results are shown in Fig~\ref{fig:imagenet-cf}. The generation results show that \IMG~is able to robustly approximate the direction of the gradient and push the image towards the decision boundary of the classifier. 
Quantitative results and human user study results for the criteria outlined in Section~\ref{subsec:eval-protoc} are shown in Table~\ref{tab:counterfactual-eval}. Because baseline methods make use of model gradients while FreeMCG does not, the flip rate of the model's decision is lower for FreeMCG than gradient-based baselines. However, FreeMCG yields by far the most human-perceived changes (\cref{tab:counterfactual-eval,fig:imagenet-cf}). In fact, FreeMCG-based counterfactuals often contain informative changes without changing the decision of the classifier while gradient-based baselines commonly produce counterfactuals that \textit{fool} the classifier to predict a different class but do not introduce semantically meaningful changes. Gradient-based methods are thus closer to \textit{adversarial attacks}, while FreeMCG better serves the purpose of counterfactual generation and produces interpretable, on-manifold changes towards the target class.

\noindent\textbf{Feature attribution.}
Feature attribution maps from representative gradient-based methods for both ImageNet and MIMIC-CXR are shown in Fig.~\ref{fig:smap}(a). ROAD evaluation results are also shown in Fig.~\ref{fig:smap}(b). We see that baseline gradient methods provide feature attributions that are similar to adversarial attacks, hence hardly interpretable. SHAP maps provide less informative patch-like artifact. \IMG, on the other hand, produces high quality and interpretable attributions that are similar to SHAP, but retaining the high-frequency details as it does not rely on arbitrary pixel clustering. In ROAD evaluation, we see that \IMG~achieves the steepest decline in model prediction, corresponding to better feature attributions, closer
to the lower-bound computed by SHAP.


\section{Conclusion}

We developed Derivative-Free Diffusion Mainfold Constrained Gradients (\IMG), an approximation of a neural network's gradient tangent to the data manifold, obtained \textit{without} access to the model weights or its real gradients. Previous explainability literature on gradient-based feature attribution and counterfactual generation have remained largely separate because the gradient could not be readily employed for counterfactual generation as it lies off the manifold and produces adversarial attacks rather than counterfactuals. With \IMG, one can obtain gradient directions on the image manifold that produce semantically meaningful changes and thus counterfactual explanations. We demonstrate through experiments that \IMG~not only enables state-of-the-art counterfactual generation but also serves as highly informative feature attributions, serving as a unified framework for both explainability methods, all without access to model weights.

{
    \small
    \bibliographystyle{ieeenat_fullname}
    \bibliography{main}
}

\clearpage
\onecolumn
\appendix
\setcounter{page}{1}
\maketitlesupplementary

\section{Proofs}
\label{sec:proof}

\proj*
\begin{proof}
    All particles $\x^{(k)}$ lie on $\Mc$ and are close to $\x$, so the vectors $\x^{(k)} - \bar\x$ approximately lie in $\Tc_\x\Mc$. For $\vb \in \Rd^d$, we have
    \begin{align*}
        \C_{\x\x} \cdot \vb &= \left( \frac{1}{K}\sum_{k=1}^K (\x^{(k)} - \bar\x)(\x^{(k)} - \bar\x)^\top \right) \cdot \vb\\
        &= \frac{1}{K}\sum_{k=1}^K (\x^{(k)} - \bar\x) \left((\x^{(k)} - \bar\x)^\top \cdot \vb \right) \\
        &= \frac{1}{K}\sum_{k=1}^K \left((\x^{(k)} - \bar\x)^\top \cdot \vb \right) (\x^{(k)} - \bar\x) \in \Tc_\x\Mc.
    \end{align*}
    Hence, for any vector $\vb \in \Rd^d$, the multiplication of $\C_{\x\x}$ on any $\vb$ is on manifold $\Mc$.
    More precisely, consider the eigen decomposition of $\C_{\x\x} = \sum_{i=1}^d \lambda_i \mathbf{e}_i \mathbf{e}_i^\top$:
    \begin{align*}
        \C_{\x\x} \cdot \nabla_{\x}\f(\x) &= \left( \sum_{i=1}^d \lambda_i \mathbf{e}_i \mathbf{e}_i^\top \right) \cdot \nabla_{\x}\f(\x)\\
        &= \sum_{i=1}^d \lambda_i \mathbf{e}_i \left( \mathbf{e}_i^\top  \cdot \nabla_{\x}\f(\x) \right)\\
        &= \sum_{i=1}^d \lambda_i \left( \mathbf{e}_i^\top  \cdot \nabla_{\x}\f(\x) \right) \mathbf{e}_i \\
        &\approx \sum_{i=1}^m \lambda_i \left( \mathbf{e}_i^\top  \cdot \nabla_{\x}\f(\x) \right) \mathbf{e}_i \in \Tc_\x \Mc
    \end{align*}
    In this approximation, the summation is limited to $m$ terms, with $m$ being the intrinsic dimension of $\Tc_\x \Mc$ (and $\Mc$).
    Thus, $\nabla_{\x}\f(\x)$ is scaled along the principal directions of the particles and contracted along the direction normal to the data manifold.
\end{proof}

\imgenkfthm*
\begin{proof}
    \begin{align}
        \C_{\x\f} &= \frac{1}{K}\sum_{k=1}^K (\x^{(k)} - \bar\x)(\f(\x^{(k)}) - \bar\f)^\top \\
        &= \frac{1}{K}\sum_{k=1}^K (\x^{(k)} - \bar\x)(\f(\x^{(k)}) - \f (\bar\x) + \f(\bar\x) - \bar\f)^\top \\
        &\stackrel{\text{(a)}}{=} \frac{1}{K}\sum_{k=1}^K (\x^{(k)} - \bar\x)\left(
        \left(\frac{\partial \f(\x)}{\partial \x}\right)_{\x = \bar\x} (\x^{(k)} - \bar\x) + \gO(\delta^2) + [\f(\bar\x) - \bar\f]
        \right)^\top \\
        &= \frac{1}{K}\sum_{k=1}^K (\x^{(k)} - \bar\x)\left(
        \left(\frac{\partial \f(\x)}{\partial \x}\right)_{\bar\x} (\x^{(k)} - \bar\x)
        \right)^\top + \gO(\delta^3) + \frac{1}{K}\sum_{k=1}^K (\x^{(k)} - \bar\x)\left(
        \underbrace{\f(\bar\x) - \bar\f}_{\rm const}
        \right)^\top \notag \\
        &\stackrel{\text{(b)}}{=} \frac{1}{K}\sum_{k=1}^K (\x^{(k)} - \bar\x)(\x^{(k)} - \bar\x)^\top
        \left(\frac{\partial \f(\x)}{\partial \x}\right)_{\bar\x}^\top + \gO(\delta^3)\\
        &= \C_{\x\x} \nabla_{\x} \f(\x)^\top + \gO(\delta^3)
    \end{align}
    where $(a)$ is given by the Taylor expansion $$\f(\x) = \f(\bar\x) + \nabla_{\x} \f(\x)|_{\bar\x}(\x - \bar\x) + \gO( \| \nabla^2 \f \| \cdot \| \x - \bar\x \|^2 )$$ and $(b)$ is from $\frac{1}{K}\sum_{k=1}^K (\x^{(k)} - \bar\x)=0$.
\end{proof}

\clearpage

\section{Chest X-ray Counterfactual Generation}
\label{sec:cxr-cf}

For CXR counterfactual generation, we empirically find that using direct gradient-ascent with FreeMCG (i.e., Eq.~\ref{eq:img_ce}), rather than reverse diffusion as was done with ImageNet, produces more realistic results. Details in~\cref{sec:implementation_details,sec:experiment}.

\subsection{CXR Counterfactuals: Disease $\rightarrow$ Normal}
\begin{figure}[H]
    \centering
    \vspace{-0.5cm}
    \includegraphics[width=0.75\linewidth]{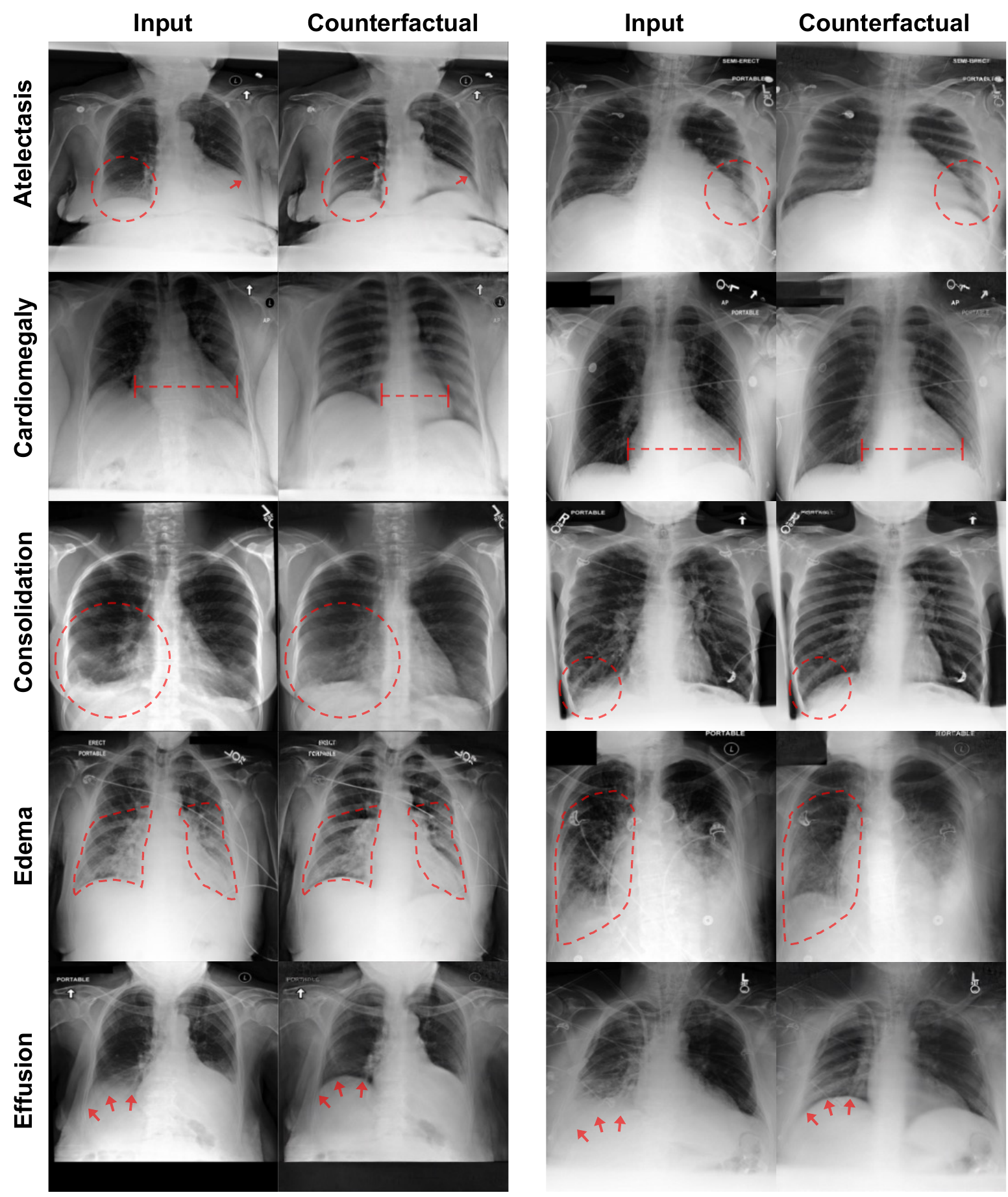}
    \vspace{0.0cm}
    \caption{Disease$\rightarrow$Normal class counterfactuals produced using FreeMCG for each type of disease lesion.}
    \label{fig:appen-cxr-cf-d2n}
    \vspace{-0.0cm}
\end{figure}

In Disease$\rightarrow$Normal counterfactuals (Fig.~\ref{fig:appen-cxr-cf-d2n}), we observe that the features associated with each lesion are removed or decreased in the generated counterfactuals. For example, the \texttt{cardiomegaly}$\rightarrow$\texttt{normal} counterfactual shows decreased heart size and \texttt{effusion} $\rightarrow$ \texttt{normal} counterfactual shows sharper costophrenic angles, corresponding to decreased appearance of effusion. Note that for cardiomegaly, the generated counterfactuals also make the ribs more distinct. This tells the human viewer that the classification model has learned to associate more visible ribs with normal CXRs without cardiomegaly. This is a type of spurious correlation arising from the distribution of training data (MIMIC-CXR) used to train the model and serves as an example of how counterfactual generation gives visual insight about the decision boundary of the model that is not obtainable from other forms of XAI techniques such as feature attribution.

\clearpage

\subsection{CXR Counterfactuals: Normal $\rightarrow$ Disease}
\begin{figure}[H]
    \centering
    \vspace{-0.5cm}
    \includegraphics[width=0.75\linewidth]{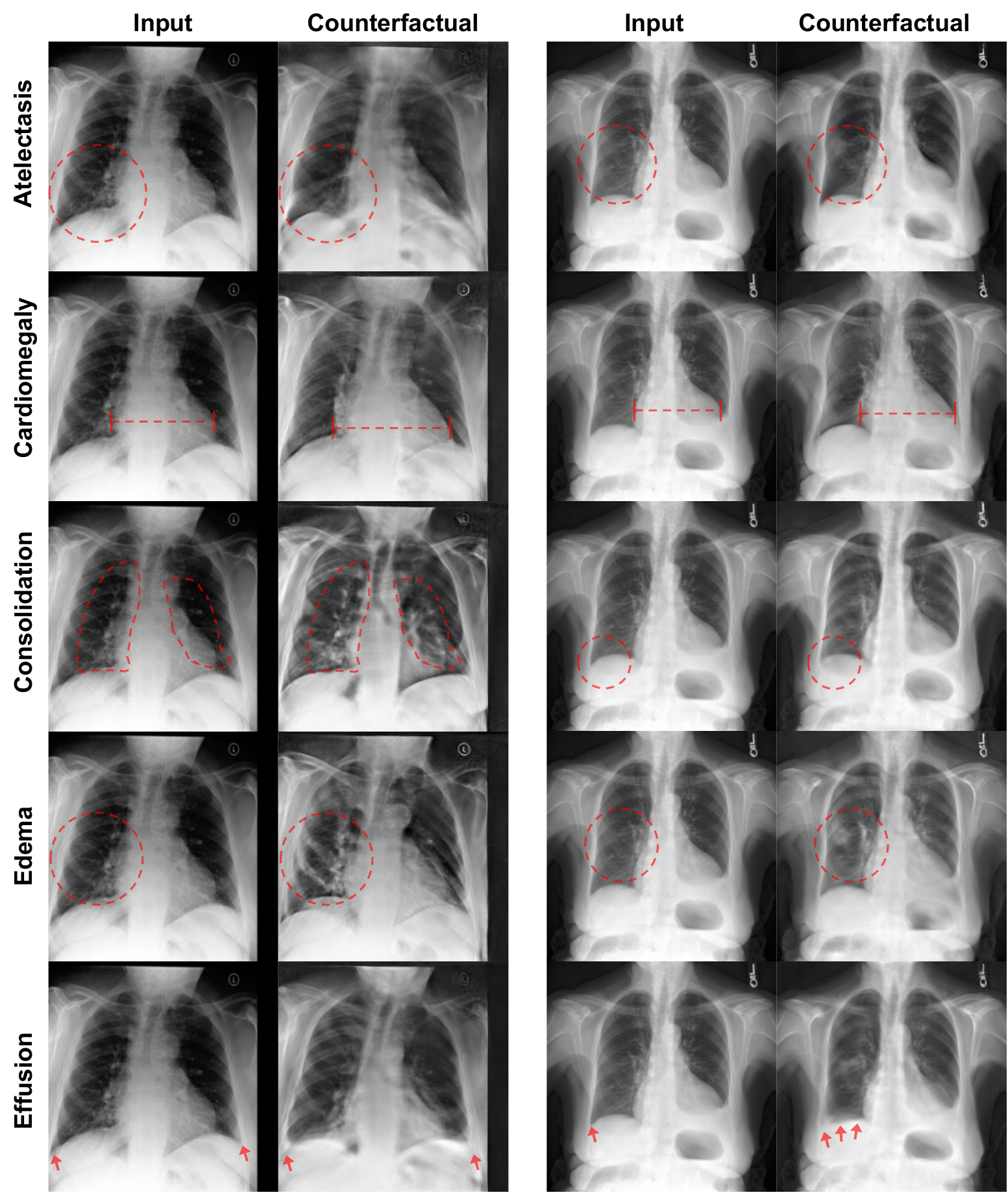}
    \vspace{-0.0cm}
    \caption{Normal$\rightarrow$Disease counterfactuals for each type of disease lesion.}
    \label{fig:appen-cxr-cf-n2d}
    \vspace{-0.0cm}
\end{figure}

From Normal$\rightarrow$Disease counterfactuals (Fig.~\ref{fig:appen-cxr-cf-n2d}), we can gain a visual understanding of what features the model associates with each lesion \eg larger cardiac diameter for cardiomegaly, blunted costophrenic angle for effusion. Note that in the example in the second column, we can see that the \texttt{normal}$\rightarrow$\texttt{consolidation} counterfactual generation elicits features more similar to effusion rather than consolidation itself. This tells us that the model associates presence of effusion with consolidation.

\clearpage

\subsection{CXR Counterfactuals: Comparison with DVCE}

Table~\ref{tab:appen-cxr-cf-eval} includes the quantitative evaluation results for CXR counterfactual generation. Because LDCE~\cite{farid2023latent} requires a separately pretrained text-to-image diffusion model, we only compare CXR counterfactual results with the DVCE~\cite{augustin2022diffusion} baseline, which is the first and prototypical method of diffusion-based counterfactual generation. We look at the Disease$\rightarrow$Normal direction counterfactual generation as we anticipate this to be the most common real usage scenario.

\begin{table*}[!thb]
    \centering
    \resizebox{0.4\textwidth}{!}{
    \begin{tabular}{lccc}
    \toprule 
      & Flip Rate $\uparrow$ & L2 $\downarrow$ & FID $\downarrow$ \\
    \midrule
    DVCE~\cite{augustin2022diffusion} & 97.9 & \textbf{526.17} & \textbf{23.57} \\
    FreeMCG & \textbf{99.2} & 1367.07 & 42.54 \\
    \bottomrule
    \end{tabular}
    
    }
    \caption{Quantitative (Flip Rate, L2, FID) and human user study (perceived change, perceived similarity, perceived realism) results.}
    \label{tab:appen-cxr-cf-eval}
    \vspace{-0.4cm}
\end{table*}

\begin{figure}[H]
    \centering
    \vspace{-0.0cm}
    \includegraphics[width=0.8\linewidth]{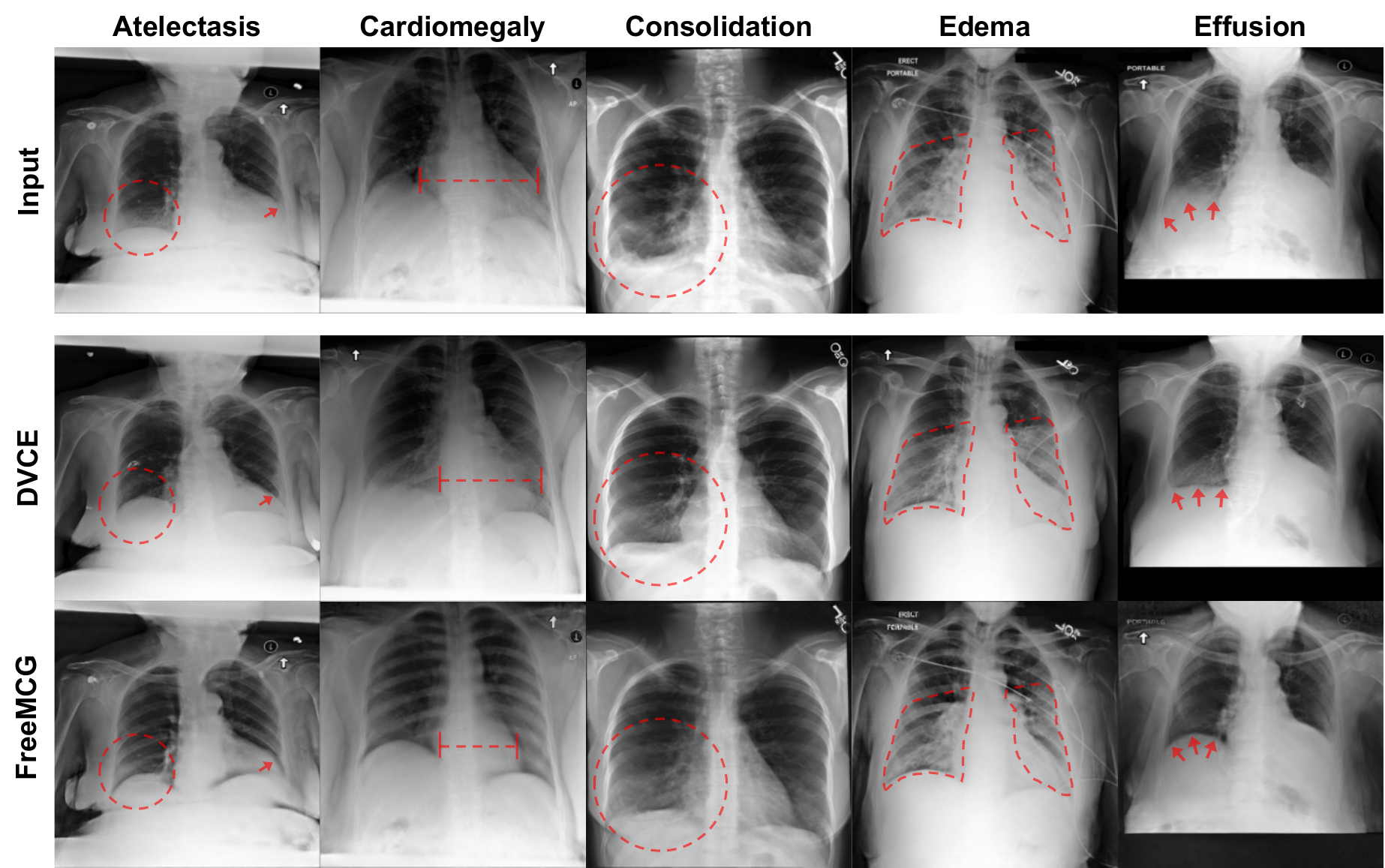}
    \vspace{-0.cm}
    \caption{Disease$\rightarrow$Normal counterfactuals for each lesion using DVCE vs. FreeMCG.}
    \label{fig:appen-cxr-cf-dvce}
    \vspace{-0.0cm}
\end{figure}

Both DVCE and FreeMCG produce counterfactuals that reduce the lesion features and move the CXR image towards the normal class, but FreeMCG produces more sparse changes than DVCE (Fig.~\ref{fig:appen-cxr-cf-dvce}). Compared to FreeMCG, DVCE more often produces outputs closer to adversarial attacks, (prediction of the classifier is flipped but changes made to the image are not very significant) resulting in lower L2 and FID. It is notable that FreeMCG, despite being a gradient-free method, achieves higher flip rate than the real gradient-based DVCE (Tab.~\ref{tab:appen-cxr-cf-eval}).

\clearpage

\section{Counterfactuals for different architectures}
\begin{figure}[H]
    \centering
    \vspace{-0.5cm}
    \includegraphics[width=\linewidth]{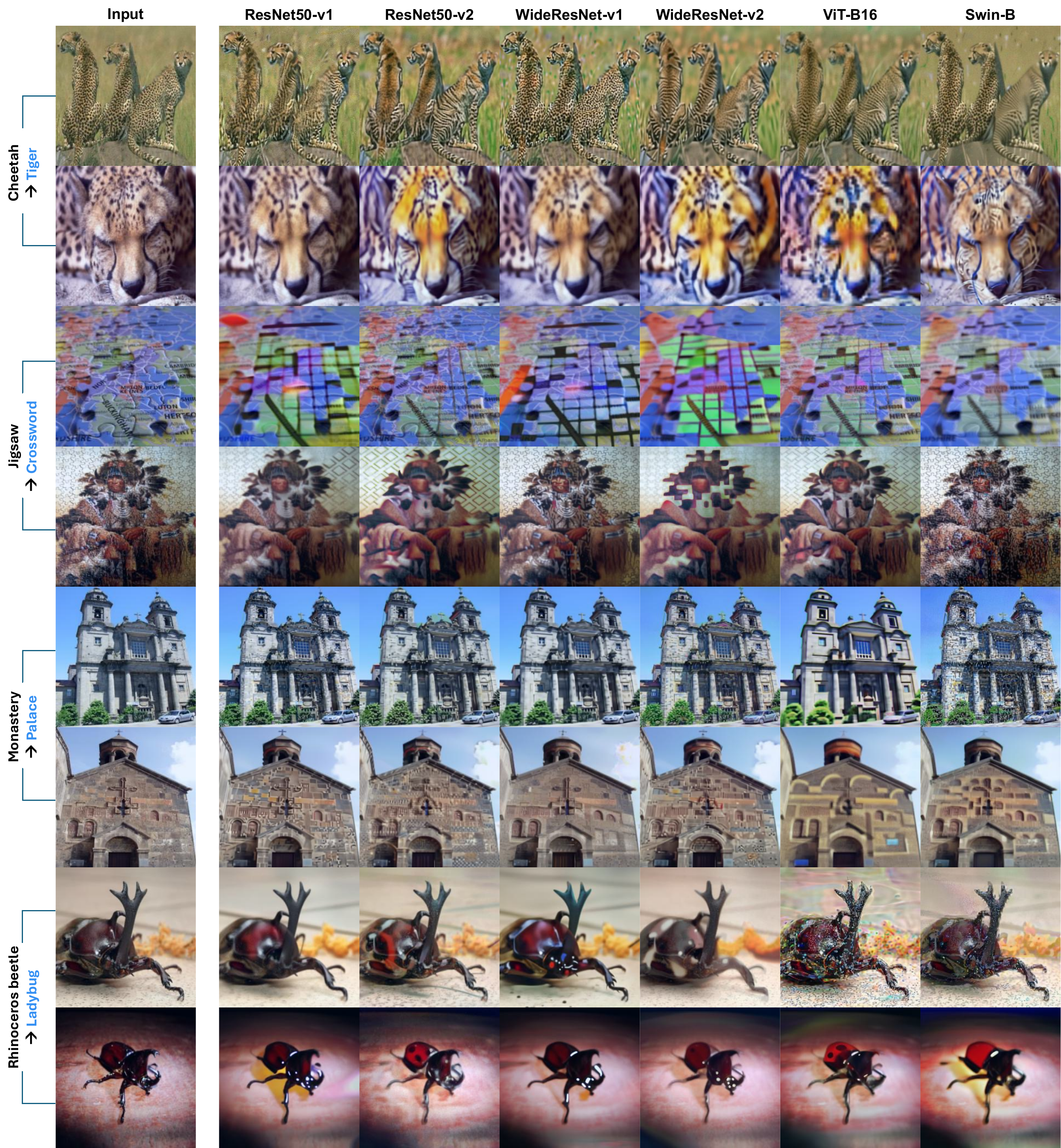}
    \vspace{-0.0cm}
    \caption{Illustration of decision boundaries of different architectures using FreeMCG-based counterfactual generation. All models are from \texttt{torchvision.models}, and \texttt{v1} and \texttt{v2} refer to different weights trained using different preprocessing pipelines. Both preprocessing and model architecture affect the decision boundary. For example, for cheetah $\rightarrow$ tiger counterfactual generation, classifiers trained using \texttt{v2} preprocessing seem to have decision boundaries more in line with human perception. In addition, certain architectures show poorly defined decision boundaries for certain input images. This type of comparison can be used to determine which models are more human-aligned for a given input image.}
    \label{fig:appen-cxr-cf-n2d}
    \vspace{-0.0cm}
\end{figure}

\section{FreeMCG Meets Latent Diffusion}

\begin{figure}[H]
    \centering
    \vspace{-0.0cm}
    \includegraphics[width=0.95\linewidth]{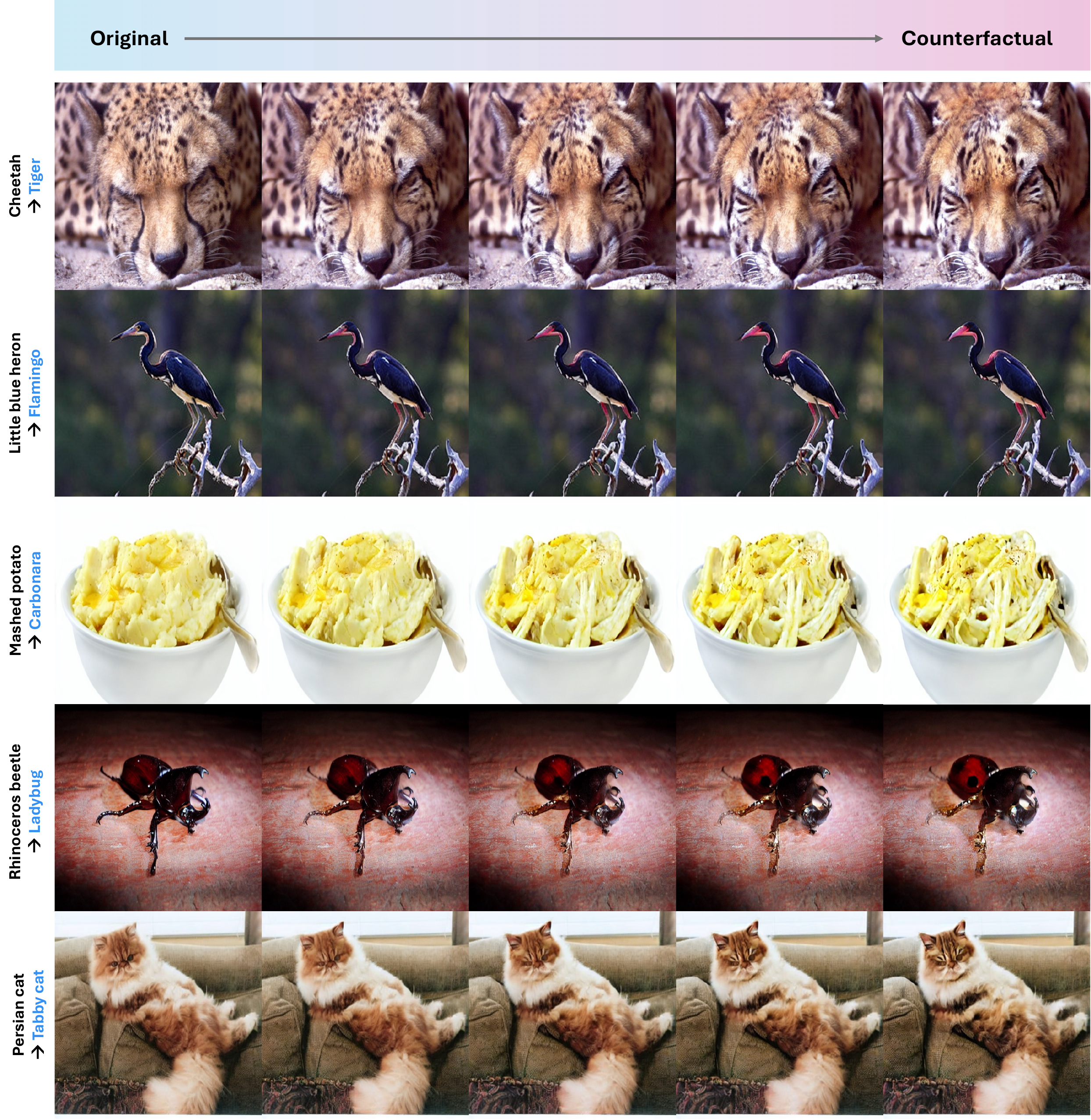}
    \vspace{-0.0cm}
    \caption{FreeMCG-counterfactual generation using Stable Diffusion~\cite{rombach2022high} with direct FreeMCG-ascent in the latent space. Leftmost column shows the input images, and rightmost column shows the final generated counterfactuals for the target class. The columns in between show intermediate steps during generation.}
    \label{fig:appen-cf-sd}
    \vspace{-0.0cm}
\end{figure}

FreeMCG can be analogously applied in the latent space to generate counterfactuals using pretrained latent diffusion models such as Stable Diffusion~\cite{rombach2022high}. Note that FreeMCG \textbf{does not use any text-conditioning} and uses only unconditional sampling. In LDCE~\cite{farid2023latent}, text conditions such as \texttt{`a photo of a tiger'} are used during generation; while this assists in the generation process, it may introduce biases of the generative model and risks reducing the faithfulness of the explanation as the generated output does not depend solely on the classification model of interest.

\clearpage

\section{Implementation details}
\label{sec:implementation_details}

For all experiments including feature attribution and counterfactual explanation, we use 100 particles per each timestep, which is much smaller than $\geq 10,000$ particles that are used in \citet{zheng2024ensemble}.

\paragraph{Feature attribution}

To save computation while taking advantage of the full scale space of the pretrained diffusion model across timesteps, we choose $t \in [100, 200, 300, 400, 500, 600, 700]$ with 100 particles per timestep to compute FreeMCG. The average across color channels is used as the feature attribution.

\paragraph{Counterfactual explanation}

For ImageNet counterfactual generation, we use the the hyperparameters $t' = 400$ and $\alpha = 0.2$. For $\beta$, we experiment with both 0.01 and 0.02 and select the most informative example. To stabilize gradients, we employ $\gamma_t := \sqrt{\bar\alpha_t \bar\alpha_{t-1}}$ as proposed in \citet{song2023pseudoinverseguided}. Further, we utilize the normalized version of the gradients as used in \citet{augustin2022diffusion}. We use DDIM~\cite{song2020denoising} sampling with 100 steps. For MIMIC-CXR counterfactual generation, we use FreeMCG directly for gradient ascent on the given input image as we find empirically that this produces more realistic results for CXR data. We use 18 iterations of gradient ascent steps at $t'=300, \alpha=0.2$.

\section{Experiment Details}
\label{sec:experiment}

\paragraph{ImageNet classifier} We use the pretrained ImageNet classification models provided by \texttt{torchvision.models}\footnote{https://pytorch.org/vision/main/models.html}. Results in the main paper are obtained using the ResNet50~\cite{he2016deep} architecture and \texttt{ResNet50\_Weights.IMAGENET1K\_V2} weights from \texttt{torchvision.models}.

\paragraph{ImageNet diffusion model} We use the pretrained unconditional diffusion model publicly available at the \texttt{openai/guided-diffusion}~\cite{dhariwal2021diffusion} repository\footnote{https://github.com/openai/guided-diffusion}. Note that one should use unconditional diffusion (rather than conditional diffusion) for counterfactual generation, as the direction of generation should depend only on the model of interest for it to be a faithful explanation of that model.

\paragraph{MIMIC-CXR classifier} We train a chest X-ray classification model on the ResNet50 architecture~\cite{he2016deep} using the MIMIC dataset~\cite{johnson2019mimic} on the five CheXpert competition labels (atelectasis, cardiomegaly, consolidation, edema, pleural effusion) along with the `normal' class.

\paragraph{MIMIC-CXR diffusion model} We train an unconditional diffusion model on MIMIC CXR images using the same recipe as the diffusion model used for ImageNet~\cite{dhariwal2021diffusion}; but because CXR images are much less diverse than ImageNet, we decrease the size of the U-net to have 128 channels and attention resolutions of 16, 8 (compared to 256 channels and attention resolutions of 32, 16, 8 for ImageNet diffusion U-net). In addition, since CXR images are greyscale images (i.e., not RGB) and therefore require only one color channel, we experimented with both 1-channel and 3-channel diffusion and found that the results were consistent with no discernible differences.

\paragraph{ROAD~\cite{rong2022consistent} for CXR feature attribution}
As there are only six classes for chest X-ray classification, rather than reporting the drop in accuracy (which were not very informative due to the fact that CXR classifiers are less accurate than ImageNet classifiers to start with and there is a relatively high chance of ``guessing'' the correct answer due to the low number of classes), we measure the drop in the predicted probability for the originally predicted class. This assessment operates on the same principle as the ROAD for ImageNet (i.e., model prediction decline with feature removal) but allows for a more fine-grained assessment.

\clearpage

\section{Human User Study}
\label{sec:human-user-study}
For the human user study, we asked each user to rate the generated counterfactuals from different methods on three different categories: \textbf{(1) change} towards target, \textbf{(2) similarity} to input, and \textbf{(3) realism} of the generated image. We present the user with the instructions, followed by each input image and corresponding counterfactuals (Fig.~\ref{fig:human-user-study}). We then prompt the user to score each counterfactual on a 5-point scale for each evaluation category.
\begin{figure}[H]
    \centering
    \vspace{-0.0cm}
    \includegraphics[width=\linewidth]{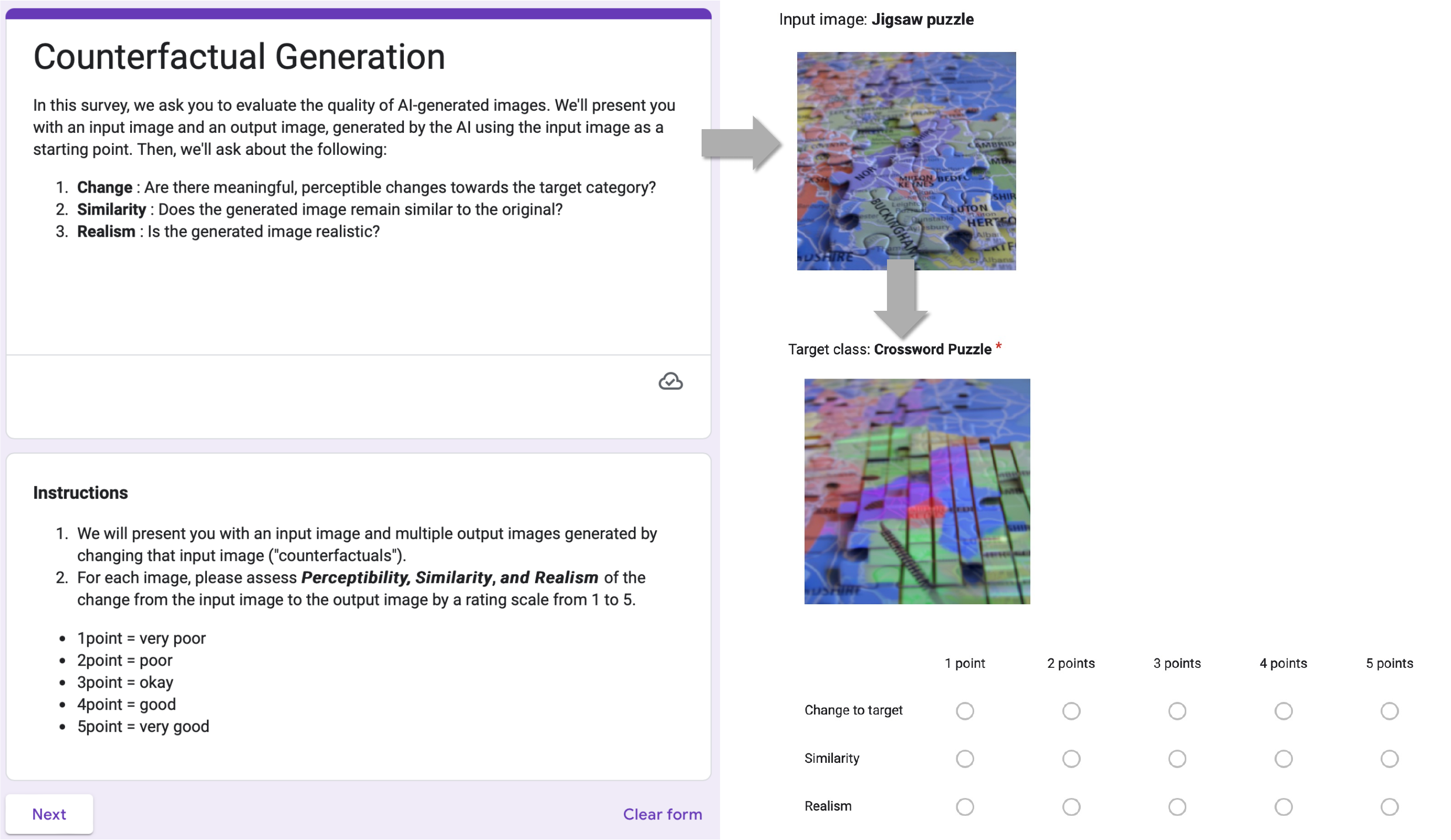}
    \vspace{-0.0cm}
    \caption{Interface for the human user study. Questions are repeated for each input image and corresponding generated counterfactuals.}
    \label{fig:human-user-study}
    \vspace{-0.3cm}
\end{figure}

\end{document}